\documentclass[journal]{IEEEtran}
\usepackage{times}
\usepackage{latexsym}
\usepackage{amsmath}
\usepackage{graphicx}
\usepackage{booktabs}
\usepackage{multirow}
\usepackage{dsfont}
\usepackage{makecell} 
\usepackage{hhline}
\usepackage{color}
\newcommand{\myRed}[1]{\textcolor{black}{#1}}

\ifCLASSINFOpdf
\else
\fi
\begin{document}
\title{Towards Textual Out-of-Domain Detection without In-Domain Labels}

\author{Di~Jin\thanks{Corresponding Author: Di Jin (djinamzn@amazon.com)},%~\IEEEmembership{Member,~IEEE,}
        ~Shuyang~Gao,%~\IEEEmembership{Fellow,~OSA,}
        ~Seokhwan~Kim,
        Yang~Liu,
        and~Dilek~Hakkani-T\"ur%~\IEEEmembership{Life~Fellow,~IEEE}% <-this % stops a space
\thanks{All authors are with Amazon Alexa AI, Sunnyvale, USA, 94089, USA.}% <-this % stops a space
% \thanks{J. Doe and J. Doe are with Anonymous University.}% <-this % stops a space
% \thanks{Manuscript received April 19, 2005; revised August 26, 2015.}
}


\markboth{}%
{Shell \MakeLowercase{\textit{et al.}}: Bare Demo of IEEEtran.cls for IEEE Journals}



\maketitle

\begin{abstract}
In many real-world settings, machine learning models need to identify user inputs that are out-of-domain (OOD) so as to avoid performing wrong actions. This work focuses on a challenging case of OOD detection, where no labels for in-domain data are accessible (e.g., no intent labels for the intent classification task). To this end, we first evaluate different language model based approaches that predict likelihood for a sequence of tokens. Furthermore, we propose a novel representation learning based method by combining unsupervised clustering and contrastive learning so that better data representations for OOD detection can be learned. Through extensive experiments, we demonstrate that this method can significantly outperform likelihood-based methods and can be even competitive to the state-of-the-art supervised approaches with label information.
\end{abstract}


% \fi
\IEEEpeerreviewmaketitle

\section{Introduction}
% for IEEE journal papers produced under \LaTeX\ using
% IEEEtran.cls version 1.8b and later.

\IEEEPARstart{D}{eep} learning models are widely used in many real-life applications, and for many classification problems. At test time, models need to identify examples that differ significantly from the model’s training data distribution, i.e., out-of-domain detection. 
For example, in dialog agents such as Amazon Alexa and Google Home Assistant, detecting  unknown  or  out-of-domain (OOD)  intents  from  user  queries  is  an  essential  component in order to  know  when  a  query  falls outside  their  range  of  predefined, supported  intents~\cite{xu2020deep}.  Correctly  identifying  out-of-scope  cases  is  especially crucial in deployed systems, both to avoid performing the wrong action and also to identify potential future directions for development. However, for state-of-the-art deep learning classifiers, it has been widely observed that their raw probability values are often over-calibrated, i.e., it has high values even for OOD inputs~\cite{guo2017calibration,hendrycks17baseline}. This necessitates having a specially designed mechanism for OOD detection. 

The task of OOD detection has historically been explored in related forms under various names such as outlier detection, anomaly detection, open classification, etc.~\cite{hendrycks17baseline,breunig2000lof,shu-etal-2017-doc,lee2018training,NEURIPS2018_abdeb6f5,jin-etal-2021-towards}. 
Most of previous work relied on the existing class labels (ID labels hereafter) for in-domain multi-class text classification tasks. For example of OOD detection for the intent classification task, existing work implement OOD detection based on a classifier that has been well trained using the intent labels for ID data~\cite{xu2020deep,podolskiy2021revisiting,zhang-etal-2020-discriminative}. 
% Most of these previous works assume class labels within in-domain (ID) data are known, therefore OOD detection is often combined with the original classification task.
In contrast, our work addresses the OOD detection problem in the scenario where no ID labels are available (e.g., no intent labels for the multi-class intent classification task) and thus most existing solutions are not directly applicable. \myRed{One real-world use case for this scenario could be the zero-shot intent classification task, which commonly happens in practice especially for early prototyping or extending a developed system to a new domain. Existing work have shown successful results without using any ID labels~\cite{xia-etal-2018-zero,namazifar2021language}, however, these studies assume that the data at test time would contain only in-domain examples. Our work can augment these systems with new capability to detect out OOD samples for real-life user data while maintaining the zero-shot capability.}
% \myRed{One real-life use case for this scenario is to detect user queries/requests that are beyond domain APIs to support frictionless task-oriented conversations, where the dialogue flow does not break when users have requests that are out of the scope of APIs/DB but potentially are already available in external knowledge sources~\cite{kim2020beyond,kim2021robust,jin-etal-2021-assistance}. This task does not assume the availability of ID labels for detecting those out-of-scope user queries.}
% ID data may contain multiple classes but we do not know how many classes there are nor do we have those annotated labels, which we call as \textit{OOD detection without ID labels}. 
% In this case, standard supervised classification with ID data cannot be performed, thus disabling the majority of current methods and imposing us a great challenge. 

Current available methods for tackling this \textit{OOD detection without ID labels} setting can be categorized into two threads: language model (LM) likelihood based methods~\cite{NEURIPS2019_1e795968,gangal2020likelihood}, and representation learning based one-class classification method~\cite{ruff-etal-2019-self,sohn2021learning}. 
For the first thread, \cite{choi2019generative,nalisnick2018do} find that likelihood is poor at separating out OOD examples; in some cases even assigning them higher likelihood than the ID test split. \cite{NEURIPS2019_1e795968} made similar observations for  detecting OOD DNA  sequences and thus proposed to “correct” the original likelihood with one from a “background” model trained on noisy inputs, termed as likelihood ratio (LR). \cite{gangal2020likelihood} explored the use of LR method on NLP tasks and reported good performance. 
For the second thread, we can treat all of the ID data as one-class data and those OOD data as other classes, forming a one-class classification problem. \cite{sohn2021learning} proposed to first learn a good representation model via contrastive learning and then  learn a density estimator on the obtained representations, which has shown state-of-the-art performance for one-class classification.

In this work, we propose novel methods for both of the above-mentioned threads for NLP tasks. 
Specifically, for the first thread of LM likelihood based methods, 
we thoroughly evaluate several likelihood based approaches, 
including that from \cite{gangal2020likelihood}.
We demonstrate that the background model trained on noisy inputs is not effective for most cases, and length normalization for LM likelihood is often needed for proper score evaluation. 
%can learn nothing but the input sequence length, which can be replaced by a simple length normalization trick that has been widely used for calculating perplexity scores~\cite{chen1998evaluation}. Through experiments, we show that the likelihood after length normalization can actually outperform or remain comparable to LLR method, which contradicts previous observations in vision and sequencing tasks~\cite{nalisnick2018do,NEURIPS2019_1e795968}. 
%Besides, we further propose to use a generic pre-trained language model such as GPT-2 as the background model, which is named as LLR+ and can significantly improve the LLR method. 
For the other direction, we find that directly applying representation learning based one-class classification methods to our problem cannot yield satisfying performance, since ID data indeed consists of multiple classes that we do not have access to. To solve this issue, we propose to perform both unsupervised clustering and contrastive learning when we learn representations of ID data, which can help produce sharper boundaries between ID and OOD samples. To the best of our knowledge, we are the first to apply the representation learning based one-class classification method to OOD detection in NLP tasks and significantly improve it so that it can rival likelihood based methods. In summary, our contributions are listed below:

\begin{itemize}
    \item We evaluate different LR methods using LM likelihood and augmenting factors including length normalization and background LMs. We demonstrate that a pre-trained background LM without any fine-tuning is even better for OOD detection than currently published background LMs trained on noisy texts. 
    \item We for the first time adopt representation learning based one-class classification method for textual OOD detection and enhance the representation learning process by introducing contrastive learning based unsupervised clustering, which plays a key role in improving detection performance for this line of methods. 
    \item We have conducted a thorough comparison between the LM likelihood based and representation learning based methods, and demonstrate that with proper unsupervised representation learning methods, the latter line of approaches can outperform the other counterpart. 
    \item Overall, our proposed methods for textual OOD detection without ID labels have achieved very impressive performance, even close to those achieved by supervised methods that use ID labels for training for three out of four datasets.\myRed{\footnote{\myRed{https://github.com/jind11/Unsupervised-OOD}}}
\end{itemize}

\section{Related Works}

\subsection{OOD Detection}

OOD detection has a long history~\cite{breunig2000lof}. 
Recent studies on  this problem include \cite{hendrycks17baseline} that proposed benchmark datasets on vision problems, and \cite{hendrycks2018deep} where OOD data were curated and used by training models to increase entropy on OOD samples. However, in real-life NLP applications, gathering and maintaining OOD data is complicated by the lexical and stylistic variation of natural language~\cite{tan-etal-2019-domain}. For this reason, methods without OOD supervision gain more attention. This is referred as \textbf{unsupervised} OOD detection, which is the focus of this work. 

All OOD detection methods for unsupervised detection rely on producing a score that can be compared with a threshold to differentiate between ID and OOD samples. One major line of methods utilize classifiers trained on the ID data to calculate the \textbf{probability} of the test instance and use it as comparison score. The maximum softmax probability is recognized as a strong baseline~\cite{hendrycks17baseline}. \cite{liang2018enhancing} found that increasing the softmax temperature $\tau$ makes the resulting probability more discriminative for OOD Detection. \cite{lee2018training} and \cite{zheng2020out} utilized ID inputs and unlabeled data to generate pseudo-OOD utterances around the decision boundaries with a generative adversarial network (GAN) so as to calibrate the output probabilities. Another important line of methods use \textbf{distance estimation} as the OOD score: \cite{NEURIPS2018_abdeb6f5} proposed using Mahalanobis distances to per-class Gaussians in the intermediate representation learned by the classifier. Specifically, a Gaussian distribution is fit for each training class from all training points in that class. Following this work, \cite{kamoi2020mahalanobis} analyzed the effectiveness of Mahalanobis distance and proposed several variants. \cite{xu-etal-2020-deep} and \cite{podolskiy2021revisiting} applied it to NLP tasks and reported strong performance. 

All of the above previous work assumed the availability of well-trained classifiers on ID data with annotated ID labels. However, in some cases, such annotated labels for classification may not be available, posing a great challenge to OOD detection. Our work will focus on addressing this special case of \textbf{OOD detection without ID labels}. It has been rarely explored previously. The only available ready-to-use method is the likelihood ratio method, where two deep generative models are trained on normal and noisy inputs, respectively, and then the difference of likelihood produced by these two models is used as the OOD score~\cite{NEURIPS2019_1e795968,gangal2020likelihood,podolskiy2021revisiting}. 
In this study, we compare that work with other ways of building the background model and evaluate the effect of length normalization for likelihood scores. 
We find that a generic pre-trained language model without domain adaptation is a better choice for the background model. 

\subsection{One-class Classification}

From another perspective, we can treat all ID data as one class and all OOD data as open classes, then we can adopt those one-class classification methods for our problem. Along this path, recent work has focused on utilizing self-supervised learning methods such as rotation learning~\cite{NEURIPS2019_a2b15837} and contrastive learning~\cite{sohn2021learning,chen2020simple} for representation learning and then adopting a density estimator to differentiate the target class from others. While we adopt this framework for textual OOD detection for the first time, we find its performance is unsatisfying. The reason is that in our problem setting, ID data are not strictly one class but indeed contain multiple hidden classes, which is not accommodated by existing one-class classification methods. 
Therefore,  we propose an unsupervised clustering method augmented by contrastive learning~\cite{zhang2021supporting,gao2021simcse} as the self-supervised method for representation learning, and show this leads to significant improvement on OOD detection.

\section{Methods} 

\myRed{Overall for OOD detection given an input text $\mathbf{x}$, we need to produce a score $s$ for it, which is used to be compared with a threshold $\eta$ to determine whether this is an ID sample or an OOD one. Such a score can be a likelihood score produced by a language model or can be a density score produced by a density estimator. We therefore propose the LM Likelihood based method for the former kind of likelihood score in Section \ref{sec:likelihood-based-method}, while proposing the representation learning based method for the density score in Section \ref{sec:representation-based-method}.}

\subsection{LM Likelihood Based}
\label{sec:likelihood-based-method}

In this section, we introduce various methods based on language models (LM).  
The basic idea using LM for OOD is building a LM using the in-domain data and then checking how likely a test instance is generated by that LM.  
The probability of an instance can be estimated using traditional n-gram based LMs. 
In this work, we adopt recent neural networks as LMs. 
%In next section, instead of calculating probabilities for the token sequence,  we first learn representations for the entire sequence, and then develop density estimator for the representation vectors. 

We evaluate the following approaches. 
\begin{itemize}
    \item LN (likelihood normalized).   
\begin{equation}
\label{eq:LLN}
    LN(\mathbf{x})=\Big{[}\prod_{i=1}^{|S|}P_{M_{ID}}(x_i|\mathbf{x}_{<i})\Big{]}^{1/|S|}
\end{equation}
where $M_{ID}$ represents the LM trained on the in-domain data, $|S|$ is the length of instance $\mathbf{x}$, $x_i$ denotes the $i^{th}$ token in $\mathbf{x}$, and $\mathbf{x}_{<i}$ denotes all tokens before $x_i$ in $\mathbf{x}$. 
This is the likelihood with length normalization using in-domain LM, and is equivalent to the widely used LM perplexity metric. 

\item LR (likelihood ratio). 
\begin{equation}
    LR(\mathbf{x})=\frac{P_{M_{ID}}(\mathbf{x})}{P_{M_B}(\mathbf{x})} = \frac{\prod_{i=1}^{|S|}P_{M_{ID}}(x_i|\mathbf{x}_{<i})}{\prod_{i=1}^{|S|}P_{M_B}(x_i|\mathbf{x}_{<i})},
\end{equation}
where $P_{M_{ID}}(\mathbf{x})$ and $P_{M_B}(\mathbf{x})$ denote the probability of $\mathbf{x}$ produced by the in-domain LM ($M_{ID}$) and a background LM ($M_B$), respectively. 

There are different ways to obtain the background LM. 
In this study, we use a generic pre-trained language model such as GPT-2 without any fine-tuning as the background LM.
% , and then fine-tune it on the in-domain samples via masked language modeling (MLM) as the ID model.

In the literature, \cite{gangal2020likelihood} also proposed using LR for OOD in NLP tasks, where they trained the in-domain model $M_{ID}$ on the original training corpus $X=\{\mathbf{x}^1,...,\mathbf{x}^N\}$ via a language modeling task, and the background model $M_{B}$ on a noisy version of training samples $\hat{X}=\{\hat{\mathbf{x}}^1,...,\hat{\mathbf{x}}^N\}$ generated with random  word  substitutions. Specifically, with probability $p_{noise}$, each  word $x_i$ in sentence $\mathbf{x}$ is randomly selected and substituted with  word $x_i'$ that is sampled  from  the vocabulary $V$ with  distribution $N(x_i')=\frac{\sqrt{f(x_i')}}{\sum_{x\in V}\sqrt{f(x)}}$, where $f(x)$ is the word frequency. 
We adopt their approach in this work for comparison. 
We call this approach $LR_{ws}$, and will show in the experiments later that such a way to create background corpus is often not appropriate. 

The difference between LR and LN is in that no background models are used in LN, and thus LN is more computationally efficient. 
Note that LR methods have been widely used in many tasks.
For example, in speaker verification or recognition, LR is calculated using the speaker specific model with respect to the universal background model.
% ~\cite{jurafsky2000speech}. 
LR was also used for vision and DNA sequencing tasks \cite{NEURIPS2019_1e795968}. 

\item NLR (length normalized likelihood ratio).
Considering that length is an important factor in LM probabilities, we also evaluate using the ratio of the normalized likelihood for OOD. Similar to LR above, we will use an off-the-shelf pre-trained language model as the background LM. NLR is defined as below: 
\begin{equation}
    NLR(\mathbf{x})= \frac{\Big{[}\prod_{i=1}^{|S|}P_{M_{ID}}(x_i|\mathbf{x}_{<i})\Big{]}^{1/|S|}}{\Big{[}\prod_{i=1}^{|S|}P_{M_B}(x_i|\mathbf{x}_{<i})\Big{]}^{1/|S|}}.
\end{equation}
\end{itemize}

\subsection{Representation Learning Based}
\label{sec:representation-based-method}

In this section, we would like to propose a representation learning based method for OOD detection without any ID labels. More specifically, the representation learning is implemented via a classic unsupervised method: unsupervised clustering. Combined with a recent popular technique, contrastive learning, it can significantly boost the OOD detection performance. 

% \subsubsection{Overall Framework}

\myRed{The training process of} our method is mainly composed of two steps: adaptive representation learning and density estimation.

\begin{itemize}
\item{\textbf{Adaptive representation learning:}} 
In this step, we learn the ID data distribution with a pre-trained sentence encoder $\psi$ by optimizing the following overall objective:

\begin{equation}
\label{eq:overall-objective}
    \mathcal{L}=\mathcal{L}_{cluster}+\gamma \mathcal{L}_{CL},
\end{equation}
where $\mathcal{L}_{cluster}$ denotes the unsupervised clustering objective and $\mathcal{L}_{CL}$ denotes the contrastive learning objective, both of which are described in more details below, and $\gamma$ is tuned on the validation set.
% $X^{ID}=\{x^{ID}_1,...,x^{ID}_N\}$ via unsupervised clustering and contrastive learning, which will be described in Section \ref{sec:clustering}. 

\item{\textbf{Density estimation:}}
In this step, we feed each ID sentence $\mathbf{x}^{ID}_i$ into the encoder $\psi$ and obtain the mean vector of all the tokens' hidden states from the last layer, which is used as its representation vector: $\psi(\mathbf{x}^{ID}_i)$. Then we learn a density estimator $D$ over these vectors, such as One-class Support Vector Machine (OC-SVM), Kernel Density Estimation (KDE), and Gaussian Mixture Model (GMM). We choose GMM in our work since we find it consistently outperforms other choices in experiments. 
\end{itemize}
%YL: consistent notation, x_i is used as a token before

\myRed{The above-mentioned training process would lead to an encoder $\psi$ and a density estimator $D$ with well-trained parameters. For inference, given a test sample $\mathbf{x}$, we first encode it with $\psi$ and then use the density estimator $D$ to produce a density score $D(\psi(\mathbf{x}))$. If this score is above a pre-set threshold $\eta$, this sample is considered as an ID sample, otherwise as an OOD sample.}

In the following, we describe the representation learning in detail. 

\subsubsection{Clustering}
\label{sec:clustering}

Our ID data may contain multiple classes, but we do not have such annotations. Therefore, we would like to adopt unsupervised clustering to learn this implicit prior. Suppose our ID data consists of $K$ semantic categories, and each category is characterized by its centroid in the representation space, denoted as $\mu_{k},k\in\{1,...,K\}$. Here $\mu_k$ is \myRed{initialized via K-means clustering and then} iteratively refined during the training phase. Note that since we do not know how many clusters there are for the ID data, $K$ is treated as a hyper-parameter and tuned using the validation set. Overall, the objective for unsupervised clustering is to push the cluster assignment probability $q_i$ for input text $x_i$ towards the target distribution $p_i$, which is achieved by optimizing the KL divergence between them: 

\begin{equation*}
    \ell^C_i=KL(p_i||q_i)=\sum_{k=1}^Kp_{ik}\log \frac{p_{ik}}{q_{ik}},
\end{equation*}
where $q_{ik}$ and $p_{ik}$ are the predicted and target probability of assigning $x_i$ to the $k^{th}$ cluster, respectively. The overall clustering objective is then defined as: 

\begin{equation}
\label{eq:clustering_loss}
\mathcal{L}_{cluster}=\frac{1}{M}\sum_{i=1}^M \ell^C_i, 
\end{equation}
where $M$ is the batch size.

Following \cite{van2008visualizing}, we use the Student’s t-distribution to compute $q_{ik}$ as below:

\begin{equation*}
    q_{ik}=\frac{(1+\parallel e_i-\mu_k\parallel^2_2/\alpha)^{-\frac{\alpha+1}{2}}}{\sum_{k'=1}^K(1+\parallel e_i-\mu_k'\parallel^2_2/\alpha)^{-\frac{\alpha+1}{2}}},
\end{equation*}
where $e_i=\psi(x_i)$, and $\alpha$ denotes the degree of freedom of the Student’s t-distribution, which is set as 1 in this work. 

Since we do not have the ground truth of $p_{ik}$, we approximate it following \cite{DBLP:conf/icml/XieGF16}:

\begin{equation*}
    p_{ik}=\frac{q_{ik}^2/f_k}{\sum_{k'=1}^Kq_{ik'}^2/f_{k'}},
\end{equation*}
where $f_k=\sum_{i=1}^Mq_{ik}$. This target distribution first sharpens the soft-assignment probability $q_{ik}$ by raising it to the second power, and then normalizes it by the associated cluster frequency. By doing so, we encourage learning from high confidence cluster assignments and simultaneously combating the bias caused by imbalanced clusters.

\subsubsection{Contrastive Learning}
\label{sec:contrastive}

Following \cite{zhang2021supporting}, we add a contrastive learning training objective~\cite{chen2020simple} while performing clustering, which can help stabilize and improve the clustering process. For a randomly sampled mini-batch $\mathcal{B}=\{x_{i^0}\}_{i^0=1}^M$ (all come from ID samples), we randomly generate an augmentation $x_{i^1}$ for each data instance $x_{i^0}$ so that $x_{i^0}$ and $x_{i^1}$ form a pair of positive instances, yielding an augmented mini-batch $\mathcal{B}^a$ of size $2M$. Following \cite{gao2021simcse}, to obtain the data augmentation, we pass the same input sentence to the pre-trained sentence encoder twice and obtain two embeddings as ``positive pairs'', by applying independently sampled dropout masks (current prevalent pre-trained encoders are all based on transformer architecture and contain dropout masks in each layer). Although strikingly simple, this approach has been found to outperform many other complex text augmentation methods, such as contextual word replacement and back-translation~\cite{gao2021simcse}. 
To achieve contrastive learning, for $x_{i^0}$ we try to bring it close to its positive counterpart $x_{i^1}$ while moving it far away from other instances in the mini-batch $\mathcal{B}^a$, which is implemented by minimizing:

\begin{equation*}
    \ell_{i^0}^{CL}=-\log \frac{\exp(sim({z}_{i^0}, {z}_{i^1})/\tau)}{\sum_{j=1}^{2M}\mathds{1}_{j\neq i^0}\cdot \exp(sim({z}_{i^0}, {z}_{j})/\tau)},
\end{equation*}
where ${z}_{j}=g(\psi({x}_{j}))$ and $g$ is implemented by a two-layers full connected network in this work; $\tau$ denotes the temperature parameter which we set as 0.5; $sim({z}_{i}, {z}_{j})={z}_{i}^T{z}_{j}/||{z}_{i}||_2||{z}_{j}||_2$ following \cite{chen2020simple}. Then the overall contrastive learning objective is defined as: 

\begin{equation}
\label{eq:CL_loss}
\mathcal{L}_{CL}=\frac{1}{2M}\sum_{i=1}^{2M} \ell^{CL}_i.
\end{equation}

% For all of the approaches above, we optimize the decision threshold on the development set.  

% Besides LSTM, we have also tried using GPT2 Small and Medium~\cite{radford2019language} as $M$ and adopt LLN as the likelihood method, whose OOD detection performance is summarized in Table \ref{tab:quant-comparison-LLR}. As can be seen, although GPT-2 is a pre-trained high-performing language model, its OOD detection performance cannot significantly outperform LSTM, indicating that the likelihood method does not benefit from large-scale pretraining for OOD detection. 

% \begin{equation}
    % \mathcal{L}_{CL}=\frac{1}{2M}\sum_{i=1}^{2M} \ell^{CL}_i.
% \end{equation}

% \paragraph{Overall objective} In summary, our overall objective is:

\section{Experiments}

% Before we introduce baselines and propose our new method, we first describe the datasets and evaluation metrics used throughout our study. 

\subsection{Datasets}

We have experimented with four text classification datasets: two on intent classification for dialogue systems, one on question topic classification, and one on search snippets topic classification. 

\paragraph{SNIPS}
consists  of user utterances for 7 intent classes such as GetWeather, RateBook, etc.~\cite{coucke2018snips}. Since this dataset itself does not include OOD utterances, we follow the procedure described in \cite{lin-xu-2019-deep} to synthetically create OOD examples. Intent classes covering at least 75\% of the training points in combination are retained as ID.  Examples  from  the  remaining  classes  are treated as OOD and removed from the training set. In the validation and test sets, examples from these classes are relabelled  to  the  single  class  label OOD. Since multiple ID-OOD splits of the classes satisfying these ratios are possible, our results are averaged across 5 randomly chosen splits.

\paragraph{ROSTD}
starts from the English part of multilingual dialog dataset released by \cite{schuster-etal-2019-cross-lingual} as ID utterances and later gets extended by \cite{gangal2020likelihood} with OOD utterances. 
% The OOD part consists mainly of subjective, under-specified, or over-emotional utterances that do not fall into ID classes.

\paragraph{Stackoverflow}
is a subset of the challenge data published by Kaggle,\footnote{https://www.kaggle.com/c/predict-closed-questions-on-stack-overflow} where question titles associated with 20 different categories are selected by \cite{xu2017self}. We follow the same way as SNIPS to create OOD samples. 
We report average results across 5 randomly chosen splits. 

\paragraph{Searchsnippets}
is extracted from web search snippets, which contains search snippets associated with 8 different topics~\cite{phan2008learning}. We randomly selected 2 out of 8 classes as OOD classes while using the remaining classes as ID classes. Again we report average results across 5 randomly chosen splits. 

Table \ref{tab:data-stats} summarizes the statistics of these four datasets. To be noted, the labels in these datasets are only used for splitting ID and OOD samples and not used for training the OOD detector.

\begin{table}[!htpb]
\centering
\small
\caption{Dataset statistics. Except ROSTD, all numbers are averaged across 5 chosen splits.}
\label{tab:data-stats}
\resizebox{0.98\columnwidth}{!}{
\begin{tabular}{lcccc}
\toprule
\textbf{Statistic} & \textbf{ROSTD} & \textbf{SNIPS} & \textbf{Stackoverflow} & \textbf{Searchsnippets} \\ \midrule
Train-ID  & 30,521 & 9,332  & 10,020         & 9,163           \\
Valid-ID  & 4,181  & 500   & 428           & 1,238           \\
Valid-OOD & 1,500  & 200   & 229           & 1,292           \\
Test-ID   & 8,621  & 506   & 418           & 1,237           \\
Test-OOD  & 3,090  & 193   & 235           & 1,293          \\ \bottomrule
\end{tabular}}
\end{table}

\subsection{Evaluation Metrics}

Following ~\cite{hendrycks17baseline}, we use the following metrics to measure OOD detection performance:

\paragraph{$\mathbf{FPR@95\%TPR}$}
corresponds to False Positive Ratio with the decision threshold being set to $\theta=sup\{\Tilde{\theta}\in \mathcal{R}|TPR(\Tilde{\theta})\leq 95\%\}$, where $TPR$ is the True Positive Ratio. Note here the OOD class is the positive class.

\paragraph{$\mathbf{AUROC}$}
measures the area under the Receiver Operating Characteristic, also known as the ROC curve. Note that this curve is for the OOD class.

\paragraph{$\mathbf{AUPR_{OOD}}$}
measures the area under Precision-Recall Curve, taking OOD as the positive class. It is more suitable for highly imbalanced data in comparison to $AUROC$.

\begin{table*}[]
\small
\centering
\caption{OOD detection performance of LM likelihood based methods on all four datasets. Since we have 5 random splits for SNIPS, Stackoverflow, and Searchsnippets datasets, we report the average and standard deviation results.}
\label{tab:quant-comparison-LLR}
\begin{tabular}{lllccc}
\toprule
\textbf{Dataset}                         & \textbf{Likelihood}           & \textbf{Model}        &  $\mathbf{AUROC} (\%)\uparrow$     & $\mathbf{AUPR_{OOD}} (\%)\uparrow$ & $\mathbf{FPR@95\%TPR} (\%)\downarrow$ \\ \midrule
\multirow{4}{*}{ROSTD}          & $LR_{ws}$                  & LSTM         & $96.81$ & $93.97$  & $13.52$ \\
                                & LN                  & LSTM         & $98.82$ & $96.91$ & $5.13$  \\ 
                                &LN & GPT-2 Medium & $98.89$ & $97.19$  & $4.74$  \\ 
                                & LR & GPT-2 Medium & $99.23$ & $97.47$  & $3.43$  \\
                                & NLR & GPT-2 Medium & $\mathbf{99.31}$  & $\mathbf{98.23}$  & $\mathbf{2.95}$  \\ \midrule
\multirow{4}{*}{SNIPS}          & $LR_{ws}$                  & LSTM         & $93.38\pm 0.82$ & $85.55\pm 2.44$  & $28.25\pm 2.63$ \\
                                & LN                  & LSTM         & $92.37\pm 0.61$  & $79.21\pm 2.23$ & $25.79\pm 2.09$  \\
                                & LN & GPT-2 Medium & $93.68\pm 2.38$ & $82.19\pm 7.79$  & $25.37\pm 9.41$  \\
                                & LR & GPT-2 Medium &  $\mathbf{95.31}\pm2.22$ & $\mathbf{88.51}\pm5.49$   & $\mathbf{21.76}\pm8.91$  \\
                                & NLR & GPT-2 Medium & $94.42\pm 2.35$ & $86.53\pm6.92$  & $25.70\pm10.77$  \\ \midrule
\multirow{4}{*}{Stackoverflow}  & $LR_{ws}$                  & LSTM         & $69.96\pm 1.03$ & $51.86\pm 1.90$ & $77.68\pm 0.46$ \\
                                & LN                  & LSTM         & $72.43\pm 0.64$ &  $57.41\pm 1.42$ & $76.86\pm 1.12$  \\
                                & LN & GPT-2 Medium & $77.76\pm 1.22$ & $60.78\pm 3.21$  & $69.12\pm 3.47$ \\ 
                                & LR & GPT-2 Medium & $\mathbf{79.77}\pm1.57$ & $60.97\pm2.31$  & $\mathbf{61.63}\pm6.44$  \\
                                & NLR & GPT-2 Medium & $78.50\pm1.79$ & $\mathbf{61.94}\pm1.88$  & $65.29\pm3.70$  \\ \midrule
\multirow{4}{*}{Searchsnippets} & $LR_{ws}$                  & LSTM         & $82.11\pm 0.75$ & $86.65\pm 0.47$ & $61.78\pm 1.16$  \\
                                & LN                  & LSTM         & $81.75\pm 2.57$ & $85.96\pm 2.69$  & $61.80\pm 1.16$  \\
                                & LN & GPT-2 Medium & $82.40\pm 1.51$ & $87.25\pm 1.67$  &  $60.12\pm 3.17$ \\ 
                                & LR & GPT-2 Medium & $80.30\pm3.25$ & $84.13\pm2.27$  & $60.67\pm7.98$  \\
                                & NLR & GPT-2 Medium & $\mathbf{82.68}\pm4.09$ & $\mathbf{87.44}\pm3.07$   & $\mathbf{59.22}\pm9.12$   \\ 
                                \bottomrule
\end{tabular}
\end{table*}

\subsection{Experimental Settings}

\subsubsection{LM Likelihood Based Methods}

For $LR_{ws}$, we follow \cite{gangal2020likelihood} and used LSTM as the LM. 
We set $p_{noise}$ to 0.5 in $LR_{ws}$, which has been found to be optimal by \cite{gangal2020likelihood}.
For $LN$, we tried both LSTM and GPT2-Medium~\cite{radford2019language} for the LM. For LSTM, we  vary  the  hidden  state dimension among $\{64, 128, 256\}$ and choose the one with highest validation set OOD detection performance. 
% To learn the LM, we use the auto-regressive language modeling objective for both LSTM and GPT2.
\myRed{To learn the in-domain LM $M_{ID}$, we train both LSTM and GPT2 on the used datasets via optimizing the unsupervised auto-regressive language modeling objective, where LSTM is trained from scratch while GPT2 is initialized with pre-trained parameters. When GPT2 is used as the background LM $M_B$, it is initialized with pre-trained parameters.}

\subsubsection{Representation Learning Based Methods}

The sentence encoder $\psi$ we used is DistilBERT-Base-NLI (denoted also as DistilBERT in the rest of this document), which is obtained by fine-tuning DistilBERT-Base on MultiNLI and SNLI datasets \cite{reimers-2019-sentence-bert}.\footnote{https://github.com/UKPLab/sentence-transformers} By tuning on the validation set, we set the number of components for GMM as 1 and set $\gamma$ as 1.0, 0.1, 1.0, and 4.0 for ROSTD, SNIPS, Stackoverflow, and Searchsnippets, respectively.

\subsection{Baselines}

We compare our representation learning based method with the following baselines:

\paragraph{Encoder w/o Fine-tuning}
We directly use a pre-trained encoder $\psi$ as an off-the-shelf model without any fine-tuning. Specifically, we used DistilBERT as well as the RoBERTa-Large-NLI-SimCSE (denoted also as RoBERTa later) that is obtained by fine-tuning RoBERTa-Large on NLI datasets via contrastive learning and is the state-of-the-art encoder for sentence embedding~\cite{gao2021simcse}.\footnote{https://github.com/princeton-nlp/SimCSE} We report the best result between these two encoders.  

\paragraph{Fine-tuned Encoder via MLM}
We fine-tune DistilBERT and RoBERTa on ID data via masked language modeling (MLM)~\cite{devlin-etal-2019-bert} to obtain domain data adapted $\psi$ and report the best result.   

\paragraph{DistilBERT/RoBERTa Maha}
We re-implemented a supervised method from \cite{podolskiy2021revisiting} by training a DistilBERT and a RoBERTa model on ID data with those ID labels and then performing OOD detection via Mahalanobis distance, which is the state-of-the-art method for textual OOD detection. 

\paragraph{$- \mathcal{L}_{cluster}$ \& $- \mathcal{L}_{CL}$}
Either unsupervised clustering by optimizing Equation \ref{eq:clustering_loss} or contrastive learning by optimizing Equation \ref{eq:CL_loss} can act as strong baselines since they can both individually serve as approaches for unsupervised representation learning. We also compare with these two baselines in the ablation study.

\section{Results}

\subsection{LM Likelihood Based Methods}
\label{sec:LLR-results}

\begin{table*}[t]
\small
\centering
\caption{Comparison of OOD detection performance for the representation learning based methods. The best results achieved by likelihood based methods from Table \ref{tab:quant-comparison-LLR} are included as one baseline (referred as Likelihood-Best). The bold font highlights the best results achieved among methods without using ID labels. $^\dag$ denotes that the best result is achieved by the RoBERTa-Large-NLI-SimCSE encoder while $^\ddag$ represents it is from DistilBERT-Base-NLI. 
\myRed{$^\ast$ denotes that the result has shown statistical significance compared with best baseline with $p<0.05$.} ``CL'' denotes ``contrastive learning''.}
\label{tab:final-performance-comparison}
\resizebox{0.95\textwidth}{!}{
\begin{tabular}{lllccc}
\toprule
\textbf{Dataset} & \textbf{ID Labels} & \textbf{Model}                        &       $\mathbf{AUROC} (\%)\uparrow$     & $\mathbf{AUPR_{OOD}} (\%)\uparrow$ & $\mathbf{FPR@95\%TPR} (\%)\downarrow$ \\ \midrule
\multirow{8}{*}{ROSTD} & \multirow{5}{*}{NO}          & Likelihood-Best                     & ${99.31}$  & ${98.23}$  & ${2.95}$  \\
                          &      & Encoder w/o Fine-tuning$^\dag$     & $\mathbf{99.77}$ & $\mathbf{99.46}$ & $\mathbf{0.71}$ \\
                          &      & Fine-tuned Encoder via MLM$^\dag$ & $99.29$ & $98.11$ & $3.00$ \\
                          &      & Ours                         & $\underline{99.73}$ & $\underline{99.30}$  & $\underline{0.73}$ \\
                          &      & \quad $- \mathcal{L}_{cluster}$ & $97.78$ & $93.91$ & $9.58$ \\
                          &      & \quad $- \mathcal{L}_{CL}$ & $98.88$ & $96.45$ & $4.87$ \\ \hhline{~-----}
\rule{0pt}{2ex}           &  \multirow{2}{*}{YES} & DistilBERT Maha & $99.61$ & $98.85$ & $1.60$  \\  
& & RoBERTa Maha     & $99.80$ & $99.50$ & $0.50$ \\ \midrule
\multirow{8}{*}{SNIPS}  &  \multirow{5}{*}{NO}      & Likelihood-Best                     & ${95.31}\pm2.22$ & ${88.51}\pm5.49$   & ${21.76}\pm8.91$  \\
                        &        & Encoder w/o Fine-tuning$^\dag$     & $94.09\pm 3.07$ & $82.31\pm 10.04$  & $17.78\pm 7.08$  \\
                        &        & Fine-tuned Encoder via MLM$^\ddag$ & $94.76\pm 3.31$ & $86.53\pm 8.07$ & $21.54\pm 15.03$ \\
                        &        & Ours                         & $\mathbf{97.56}^\ast\pm 1.01$ & $\mathbf{92.80}^\ast\pm 3.19$  & $\mathbf{8.53}^\ast\pm 4.42$  \\ 
                        &        & \quad $- \mathcal{L}_{cluster}$ & $84.54\pm 8.95$ & $64.02\pm 18.70$  & $43.57\pm 19.08$  \\
                        &        & \quad $- \mathcal{L}_{CL}$ & $97.17\pm 1.37$ & $92.22\pm 3.97$ & $15.75\pm 8.93$ \\ \hhline{~-----}
 \rule{0pt}{2ex}        &  \multirow{2}{*}{YES}      & DistilBERT Maha & $97.93\pm1.37$ & $95.29\pm3.03$ & $8.65\pm6.42$ \\
 & & RoBERTa Maha     & $98.30\pm 0.98$ & $95.10\pm 3.80$ & $7.06\pm 3.51$  \\ \midrule
\multirow{8}{*}{Stackoverflow} & \multirow{5}{*}{NO}  & Likelihood-Best                     & ${79.77}\pm1.57$ & $60.97\pm2.31$  & ${61.63}\pm6.44$ \\
                            &    & Encoder w/o Fine-tuning$^\ddag$     & $73.07\pm 3.77$ & $55.69\pm 3.40$  & $76.54\pm 1.62$  \\
                            &    & Fine-tuned Encoder via MLM$^\ddag$ & $78.59\pm 1.68$ & $62.18\pm 1.63$  & $64.13\pm 5.04$  \\
                            &    & Ours                         & $\mathbf{83.65}^\ast\pm 0.13$ & $\mathbf{62.22}\pm 2.85$ & $\mathbf{36.82}^\ast\pm 1.84$ \\ 
                            &    & \quad $- \mathcal{L}_{cluster}$ & $70.99\pm 1.99$ & $53.76\pm 3.00$  & $78.43\pm 3.32$  \\
                            &    & \quad $- \mathcal{L}_{CL}$ & $50.79\pm 13.13$ & $37.99\pm 13.54$ & $90.80\pm 5.24$ \\ \hhline{~-----}
 \rule{0pt}{2ex}    &  \multirow{2}{*}{YES}      & DistilBERT Maha & $92.00\pm0.96$ & $78.35\pm3.24$ & $19.99\pm3.02$ \\
 & & RoBERTa Maha     & $91.98\pm 1.34$  &  $76.55\pm 5.62$ & $19.61\pm 2.51$  \\ \midrule
\multirow{8}{*}{Searchsnippets} & \multirow{5}{*}{NO} & Likelihood-Best                     & ${82.68}\pm4.09$ & ${87.44}\pm3.07$   & ${59.22}\pm9.12$\\
                            &    & Encoder w/o Fine-tuning$^\dag$     & $82.93\pm 5.05$ & $88.22\pm 3.78$  & $61.80\pm11.02$ \\
                            &    & Fine-tuned Encoder via MLM$^\dag$ & $84.32\pm 1.96$ & $89.27\pm 2.15$  & $62.54\pm 4.24$  \\
                            &    & Ours                         & $\mathbf{94.70}^\ast\pm 2.29$ & $\mathbf{96.16}^\ast\pm 1.92$  & $\mathbf{22.43}^\ast\pm 7.20$ \\
                            &    & \quad $- \mathcal{L}_{cluster}$ & $88.09\pm 6.63$ & $90.91\pm 5.73$  & $41.09\pm 17.76$  \\
                            &    & \quad $- \mathcal{L}_{CL}$ & $78.95\pm 4.93$ & $82.01\pm 4.22$ & $65.06\pm 11.53$ \\ \hhline{~-----}
\rule{0pt}{2ex}             & \multirow{2}{*}{YES}      & DistilBERT Maha & $95.09\pm2.43$ & $96.88\pm1.56$ & $25.13\pm14.00$ \\
& & RoBERTa Maha     & $97.26\pm 1.05$ & $98.38\pm 0.64$ & $14.61\pm 7.00$  \\ 
\bottomrule
\end{tabular}}
\end{table*}

We first present results for the LM likelihood based methods in Table \ref{tab:quant-comparison-LLR}. 
We can see that overall LR and NLR with GPT-2 as the background LM achieve the best results for all the datasets. 
The benefit of length normalization in NLR compared with LR is not consistent across the four datasets, which we hypothesize is because the likelihood is already normalized using the background LM in the LR method. 
In contrast, for LN, since it only uses the LM likelihood from the in-domain model, the likelihood score needs to be properly normalized by the sequence length (given how the chain rule is used in probability calculation). Such a simple normalization strategy by using the sequence length can avoid using a second LM, and thus it is computationally more efficient, while its performance is actually comparable to both LR and NLR across all datasets. 
% LN is similar to using a background LM that assigns the same probability to every token. 
The comparison of LSTM and GPT2 as the in-domain LM for LN shows that GPT2 is a more powerful LM and has better quality in its likelihood estimation. 

Finally, $LR_{ws}$ is not competitive in most cases compared to LN, both of which use LSTM as their LMs. It indicates that substituting randomly selected words with other words randomly drawn from the vocabulary to create noisy texts for training the background LM is not effective. Intuitively, training a LM on perturbed texts that are totally not grammatical can learn nothing in linguistics but the sentence length, considering that the likelihood of the whole sequence is the product of all token probabilities and each token probability would be biased to be uniformly distributed over the vocabulary when learned on randomly perturbed text. This can help explain why $LR_{ws}$ cannot even win over LN in most cases since LN can also remove the influence of sentence length via normalization. We provide a more detailed analysis over $LR_{ws}$ to explain the disadvantages of this method in the Section \ref{sec:analysis_of_LR_ws}.

The large performance variation of the same methods across different datasets reveals the diverse difficulty levels of OOD detection across datasets, and the difficulty level depends on both the language variation within ID data and the overlap of data distributions between ID and OOD data. For example, in the ROSTD dataset, models achieve almost perfect OOD detection performance. This is because its language variation in the samples for each intent is very limited (e.g., many utterances for the same intent only differ in one or two tokens), and its OOD samples differ significantly from the ID ones (e.g., consisting of subjective and emotional utterances), leading to lesser overlap between ID and OOD data than other datasets. 

\subsection{Representation Learning Based Methods}

Table \ref{tab:final-performance-comparison} summarizes the main results of OOD detection for the representation learning based methods. We have included the best results achieved by those likelihood based methods from Table \ref{tab:quant-comparison-LLR} as one baseline for comparison. From this table, we have the following observations:

\begin{itemize}
    \item Simply applying a generic pre-trained encoder without any fine-tuning on the ID data to obtain data representations and then performing density estimation can already achieve good performance (refer to Encoder w/o Fine-tuning), sometimes comparable to the best likelihood method (refer to Likelihood-Best).  
    \item Adapting the pre-trained encoder on the ID data via MLM, i.e., fine-tuned encoder via MLM,  shows some improvement for most datasets and metrics, e.g., on SNIPS, Stackoverflow, and Searchsnippets datasets for the $AUROC$ and $AUPR_{OOD}$ metrics.  
    \item Our proposed method for adapting the encoder via clustering and contrastive learning can significantly boost the OOD detection performance consistently for all datasets, outperforming all baselines.  
    \item Representation learning based methods can outperform likelihood based methods significantly on all datasets. For example of the Searchsnippets dataset, the performance gap between these two lines of methods is over 10\% for all three metrics.
    \item Our method can be comparable to the state-of-the-art method that intensively utilizes the ID labels for supervised training (i.e., DistilBERT/RoBERTa Maha) for ROSTD, SNIPS, and Searchsnippets datasets. 
\end{itemize}

\myRed{The key to the success of OOD detection is to learn precise boundaries of ID data distribution, which are used to differentiate ID and OOD samples. Likelihood based methods via language modeling treat all in-domain data points equally and learn their distribution boundaries without explicitly exploring their internal distribution patterns. However, the representation based method can better make use of the fact that there exist some clusters for ID samples but we just do not have labels for them. The learned boundaries by taking into account internal clusters among ID samples can be more tight that those by treating all ID data as a single cluster, which could be the main reason why representation learning based methods significantly outperform likelihood based methods.}

It is worth pointing out that our method is based on a small-size encoder, i.e., DistilBERT-Base (around 66M parameters), however, those baselines including Encoder w/o  Fine-tuning, Finetuned Encoder, and RoBERTa Maha have adopted an encoder of much larger size, i.e., RoBERTa-Large with around 355M parameters. Moreover, the RoBERTa-Large encoder has been pre-trained on around 10 times larger corpora compared with DistilBERT-Base, thus achieving much better performance on various kinds of language understanding tasks, e.g., the average score of 88.5 achieved by RoBERTa-Large vs 77.0 achieved by DistilBERT-Base on the development sets of GLUE benchmark~\cite{Sanh2019DistilBERTAD,Liu2019RoBERTaAR}.
Although our method is equipped with an encoder with much smaller model size (about 1/5) and inferior down-stream tasks performance, it can still beat those baselines using larger models when no ID labels are used, and be on par with the baseline that does use ID labels. This comparison demonstrates that our method can achieve both high OOD detection performance and parameter \& computation efficiency, which fulfills the two consideratas for real-world deployment in industry applications.  

\myRed{Among all datasets, it can be observed that the advantage of our method over baselines for the Stackoverflow dataset is not as large as other datasets. This is mainly because this dataset is more challenging and the unsupervised clustering performance is not good on it. On one hand, this task aims to classify a short question title (usually around 10 tokens) from the Stackoverflow website into one technological topic, e.g., excel, oracle, bash, apache, etc., which is challenging in its nature. On the other hand, the best clustering algorithm in \cite{xu2017self} can only achieve 51.14\% of accuracy and the SOTA OOD model with supervised ID labels in our Table \ref{tab:final-performance-comparison} (i.e., DistilBERT Maha) can only obtain 78.35\% of $AUPR_{OOD}$, which further validates its difficulty level.}

\section{Discussions}

\subsection{Analysis of $LR_{ws}$}
\label{sec:analysis_of_LR_ws}

\begin{figure}[!hptb]
\flushleft
\includegraphics[width=1\columnwidth]{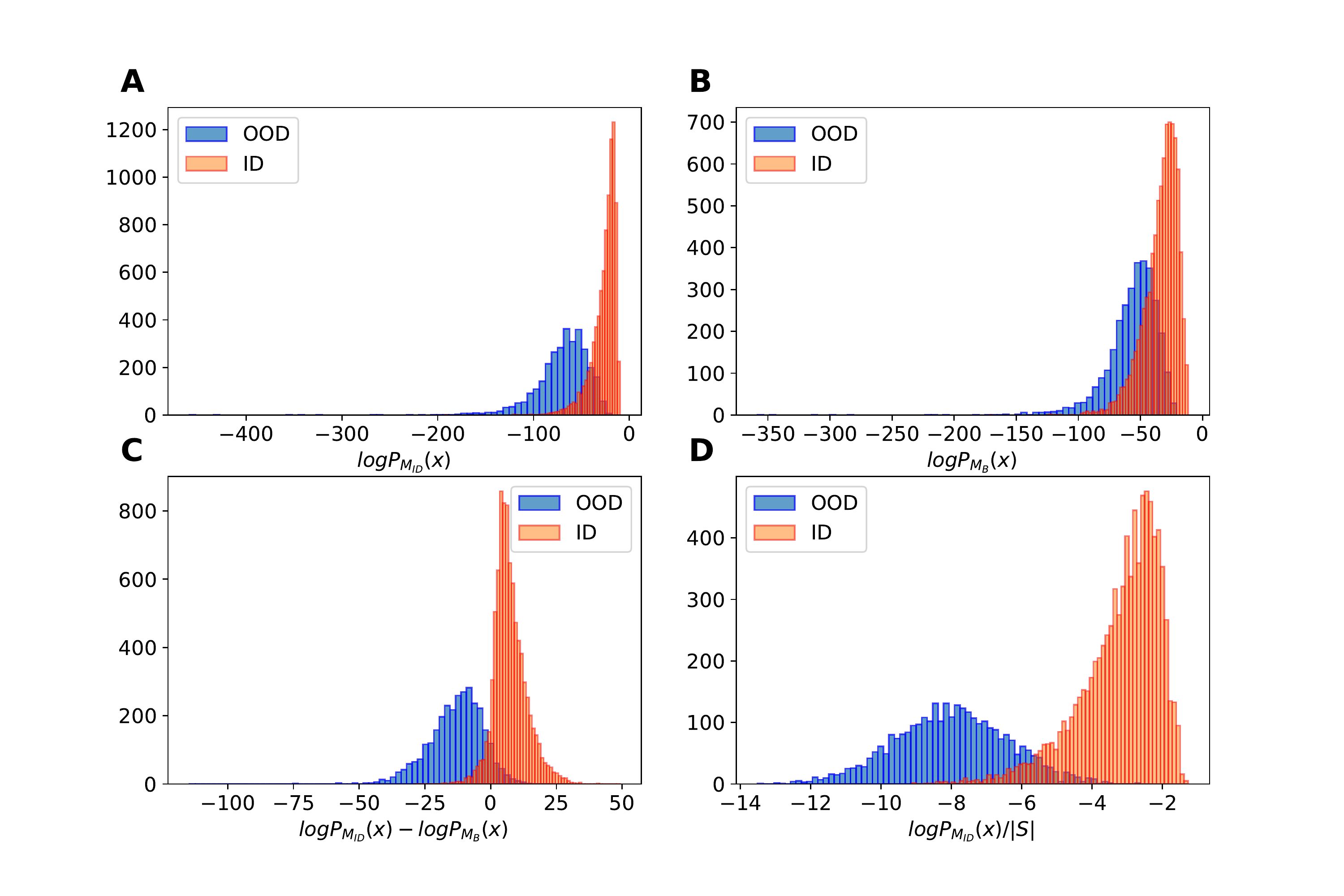}
\caption{Histogram of four kinds of likelihood for the ROSTD dataset \myRed{with LSTM as the base model}. A: $\log P_{M_{ID}}(x)$; B: $\log P_{M_B}(x)$; C: $\log LR_{ws}$; D: $\log LN$.}
\label{fig:LLR-hist-rostd}
\end{figure}

In order to further analyze the method $LR_{ws}$ reported in \cite{gangal2020likelihood}, we first plot histograms of $\log P_{M_{ID}}(\mathbf{x})$, $\log P_{M_B}(\mathbf{x})$, $\log LR_{ws}(\mathbf{x})$, and $\log LN(\mathbf{x})$ for the ROSTD dataset using LSTM as the LM, as shown in Figure \ref{fig:LLR-hist-rostd}. 
% Similar histograms for the other datasets of SNIPS, Stackoverflow, and Searchsnippets are shown in Figure \ref{fig:LLR-hist-snips}, \ref{fig:LLR-hist-stackoverflow}, and \ref{fig:LLR-hist-searchsnippets}, respectively. 
As can be seen from Figure \ref{fig:LLR-hist-rostd}B, the likelihood $P_{M_B}(\mathbf{x})$ computed by the background model $M_B$ itself, which is trained on text with noise introduced, cannot differentiate between ID and OOD samples. The overlap between histograms of ID and OOD samples of $P_{M_B}(\mathbf{x})$ (Figure \ref{fig:LLR-hist-rostd}B) is even larger than that of $P_{M_{ID}}(\mathbf{x})$ (Figure \ref{fig:LLR-hist-rostd}A). However, their ratio, $LR_{ws}$, presents less overlap between ID and OOD samples (Figure \ref{fig:LLR-hist-rostd}C). To explore the reason why subtracting the likelihood of background model can help differentiate ID and OOD samples, 
we further calculate the Pearson correlation between the likelihoods and sentence length for all four datasets, which is summarized in Table \ref{tab:LLR-correlation}. From this table, we find that $P_{M_B}(\mathbf{x})$ has a strong correlation with sentence length for all four datasets, indicating that it carries abundant information about sentence length. In contrast, $LR_{ws}$ presents much less correlation with the sentence length, which also shows less overlap between histograms of ID and OOD samples as shown in Figure \ref{fig:LLR-hist-rostd}C. This finding leads to our hypothesis that the key to the success of $LR_{ws}$ could be that the likelihood of the background model can help remove the sentence length information from $P_{M_{ID}}(\mathbf{x})$. To validate this hypothesis, we directly remove the sentence length information from $\log P_{M_{ID}}(\mathbf{x})$ by dividing it with the sentence length $|S|$, which forms $\log LN$. As shown in Table \ref{tab:LLR-correlation}, LN shows very small correlation with the sentence length and its histogram in Figure \ref{fig:LLR-hist-rostd}D shows even less overlap between ID and OOD samples compared with $LR_{ws}$. This comparison explains the two advantages of LN over $LR_{ws}$, i.e., its comparable or even superior OOD detection performance and less computation and complexity.

% \begin{figure}[!hptb]
% \flushleft
% \includegraphics[width=1\columnwidth]{figs/LLR_hist_snips.pdf}
% \caption{Histogram of four kinds of likelihood for the SNIPS dataset. A: $\log P_{ID}(x)$; B: $\log P_B(x)$; C: $\log LLR_{ws}$; D: $\log LLN$.}
% \label{fig:LLR-hist-snips}
% \end{figure}

% \begin{figure}[!hptb]
% \flushleft
% \includegraphics[width=1\columnwidth]{figs/LLR_hist_stackoverflow.pdf}
% \caption{Histogram of four kinds of likelihood for the Stackoverflow dataset. A: $\log P_{ID}(x)$; B: $\log P_B(x)$; C: $\log LLR_{ws}$; D: $\log LLN$.}
% \label{fig:LLR-hist-stackoverflow}
% \end{figure}

% \begin{figure}[!hptb]
% \flushleft
% \includegraphics[width=1\columnwidth]{figs/LLR_hist_searchsnippets.pdf}
% \caption{Histogram of four kinds of likelihood for the Searchsnippets dataset. A: $\log P_{ID}(x)$; B: $\log P_B(x)$; C: $\log LLR_{ws}$; D: $\log LLN$.}
% \label{fig:LLR-hist-searchsnippets}
% \end{figure}

\begin{table*}[t]
\centering
\small
\caption{Pearson correlation coefficient and p-value between sentence length and likelihood for all four datasets.}
\label{tab:LLR-correlation}
\resizebox{0.85\textwidth}{!}{
\begin{tabular}{lcccccccc}
\toprule
 & \multicolumn{2}{c}{\textbf{ROSTD}}                             & \multicolumn{2}{c}{\textbf{SNIPS}}                                 & \multicolumn{2}{c}{\textbf{Stackoverflow}}                           & \multicolumn{2}{c}{\textbf{Searchsnippets}}                         \\
 & \textbf{r}                         & \textbf{p-value}                   & \textbf{r}                          & \textbf{p-value}                      & \textbf{r}                          & \textbf{p-value}                        & \textbf{r}                          & \textbf{p-value}                       \\ \midrule
$P_{ID}(x)$ & -0.566                    & 0.000                     & -0.732                     & $3.279\times10^{-118}$                   & -0.803                     & 0.000                          & -0.827                     & 0.000                         \\
$P_B(x)$ & \textbf{-0.785}                    & 0.000                     & \textbf{-0.866}                     & $1.378\times10^{-212}$                   & \textbf{-0.945}                     & 0.000                          & \textbf{-0.950}                     & 0.000                         \\
$LR_{ws}$ & 0.014       & 0.122       & -0.225    & $1.762\times10^{-9}$      & -0.570        & $1.913\times10^{-172}$        & -0.176         & $3.080\times10^{-39}$ \\
LN & -0.026                    & 0.005                     & -0.099                     & 0.008                        & 0.044                      & 0.049                          & -0.053                     & $8.461\times10^{-5}$                     \\ \bottomrule
\end{tabular}}
\end{table*}

\subsection{Ablation Study}
\label{sec:ablation-study}

We did an ablation study by removing either $\mathcal{L}_{cluster}$ or $\mathcal{L}_{CL}$ when optimizing Equation \ref{eq:overall-objective} for the representation learning based methods.
These results are included in Table \ref{tab:final-performance-comparison}. By comparing with the full solution, we see that: (1) Both training objectives contribute to the improvement. (2) In some cases (e.g., Stackoverflow and Searchsnippets datasets), removing $\mathcal{L}_{CL}$ would cause larger performance degradation, while in the other cases, $\mathcal{L}_{cluster}$ is more important. 
This is dependent on the distributions of the ID samples (separation of different classes) and how OOD data is separated from ID data.
We will show visualizations analysis of data later.

\subsection{Influence of $K$}

Since we do not have any information on the ID labels, we need to tune the number of clusters $K$ \myRed{by optimizing the OOD detection performance metric $AUPR_{OOD}$ on the validation set}.
Table \ref{tab:K-tuning} shows these values for different data sets along with their numbers of classes for the ID data  for the representation learning based methods. We can see that the optimal number of clusters for OOD detection is not equal to the actual number of labels within ID data. 
For example, the optimal $K$ is 30 for the Searchsnippets dataset, which is far away from the number of labels, i.e., 6. This implies that our method is not simply an unsupervised approach to learning the classification task but rather a method more suitable for learning ID data distribution. 

%YL:  anything to say about the numbers or difference in relation to the performance on different datasets? 

\begin{table}[!htpb]
\centering
\small
\caption{Comparison between the optimal number of clusters and the number of ID labels. Since three datasets have 5 random splits, we list both minimum and maximum number of labels within the parentheses.}
\label{tab:K-tuning}
\resizebox{0.98\columnwidth}{!}{
\begin{tabular}{lcccc}
\toprule
 & \textbf{ROSTD} & \textbf{SNIPS} & \textbf{Stackoverflow} & \textbf{Searchsnippets} \\ \midrule
Optimal $K$  & 14 & 6  & 17         & 30           \\
Num of labels  & 12  & (5, 5)   & (13, 14)           & (6, 6)          \\ \bottomrule
\end{tabular}}
\end{table}

\subsection{Visualization of Representations}

\begin{figure}[t]
\centering
\includegraphics[width=0.8\columnwidth]{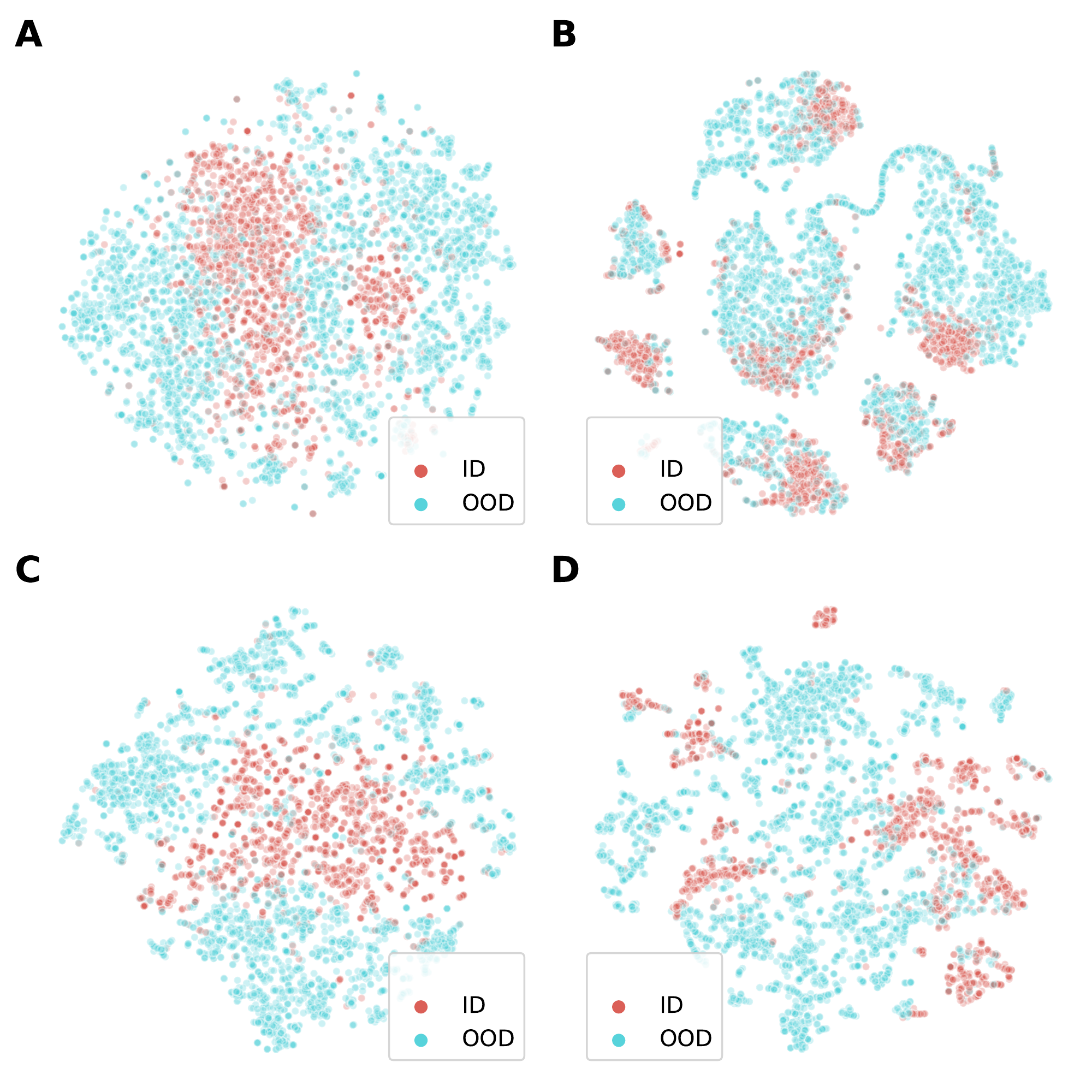}
\caption{T-SNE visualization of learned sentence representations for both ID and OOD samples of Searchsnippets dataset. A: DistilBERT-Base-NLI; B: DistilBERT-Base-NLI fine-tuned via unsupervised clustering; C: DistilBERT-Base-NLI fine-tuned via contrastive learning; D: DistilBERT-Base-NLI fine-tuned via combining clustering and contrastive learning.}
\label{fig:representation-searchsnippets}
\end{figure}

\begin{figure}[t]
\centering
\includegraphics[width=0.8\columnwidth]{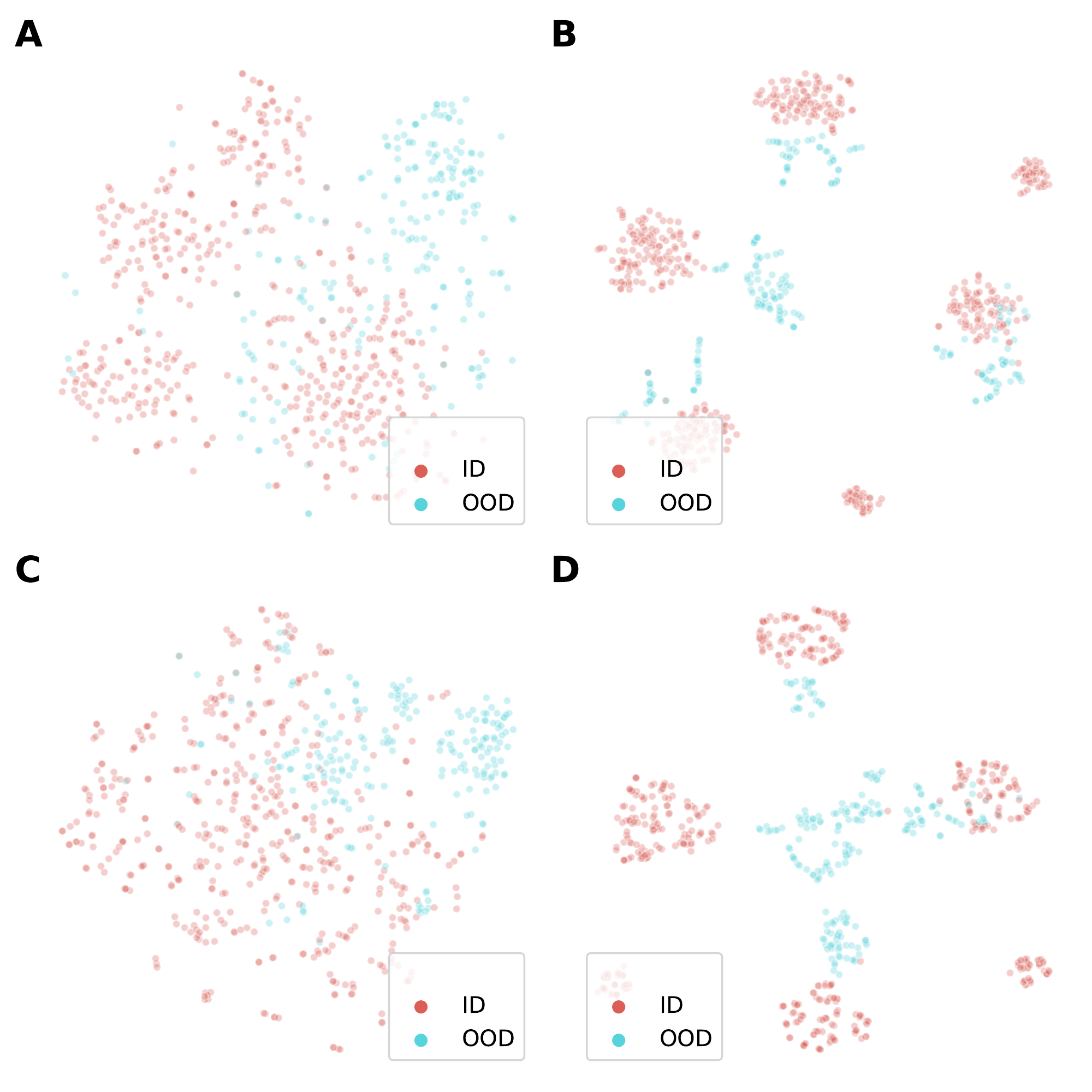}
\caption{T-SNE visualization of learned sentence representations for both ID and OOD samples of SNIPS dataset. A: DistilBERT-Base-NLI; B: DistilBERT-Base-NLI fine-tuned via unsupervised clustering; C: DistilBERT-Base-NLI fine-tuned via contrastive learning; D: DistilBERT-Base-NLI fine-tuned via combining clustering and contrastive learning.}
\label{fig:representation-snips}
\end{figure}

\begin{figure}[t]
\centering
\includegraphics[width=0.8\columnwidth]{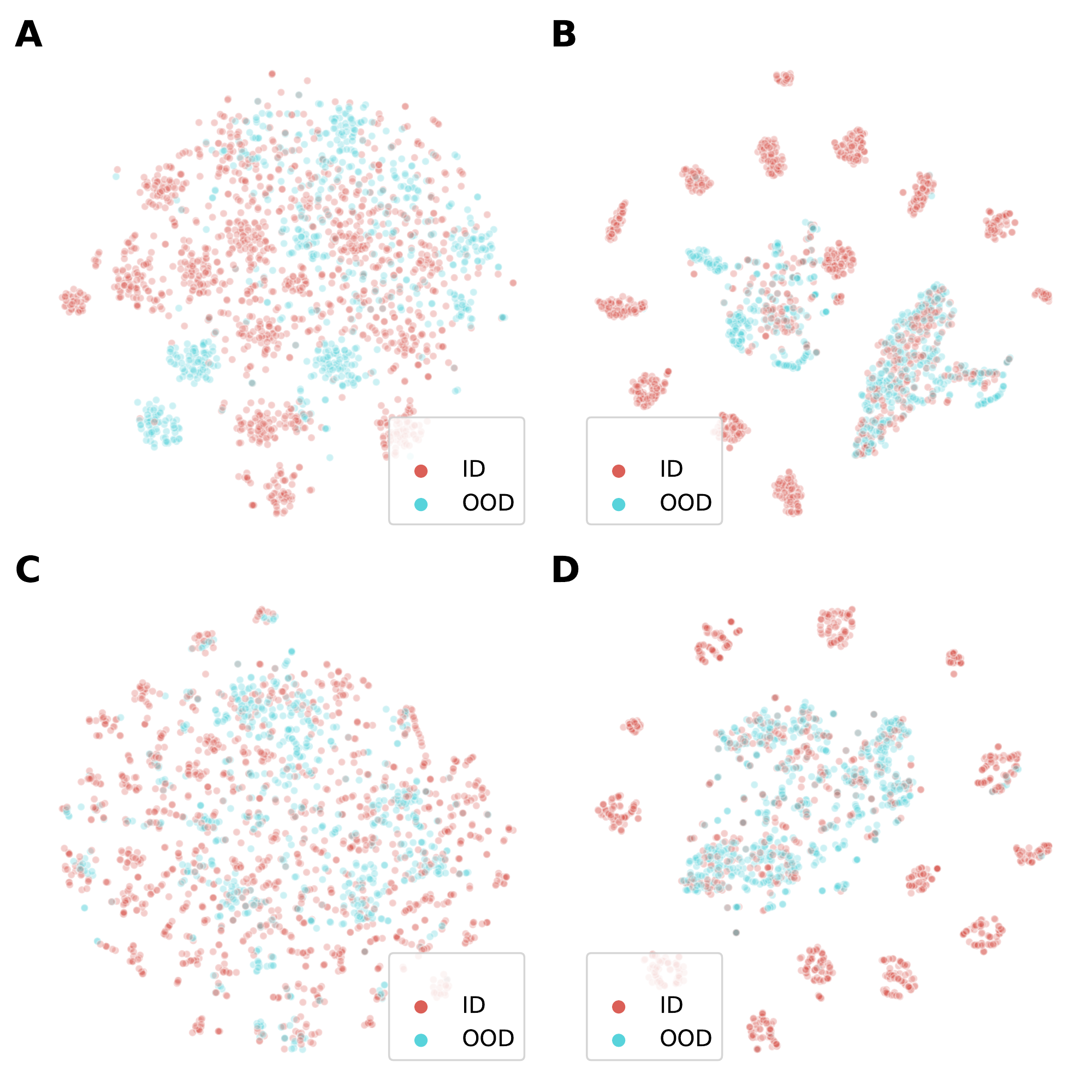}
\caption{T-SNE visualization of learned sentence representations for both ID and OOD samples of Stackoverflow dataset. A: DistilBERT-Base-NLI; B: DistilBERT-Base-NLI fine-tuned via unsupervised clustering; C: DistilBERT-Base-NLI fine-tuned via contrastive learning; D: DistilBERT-Base-NLI fine-tuned via combining clustering and contrastive learning.}
\label{fig:representation-stackoverflow}
\end{figure}

\begin{figure}[ht]
\centering
\includegraphics[width=0.8\columnwidth]{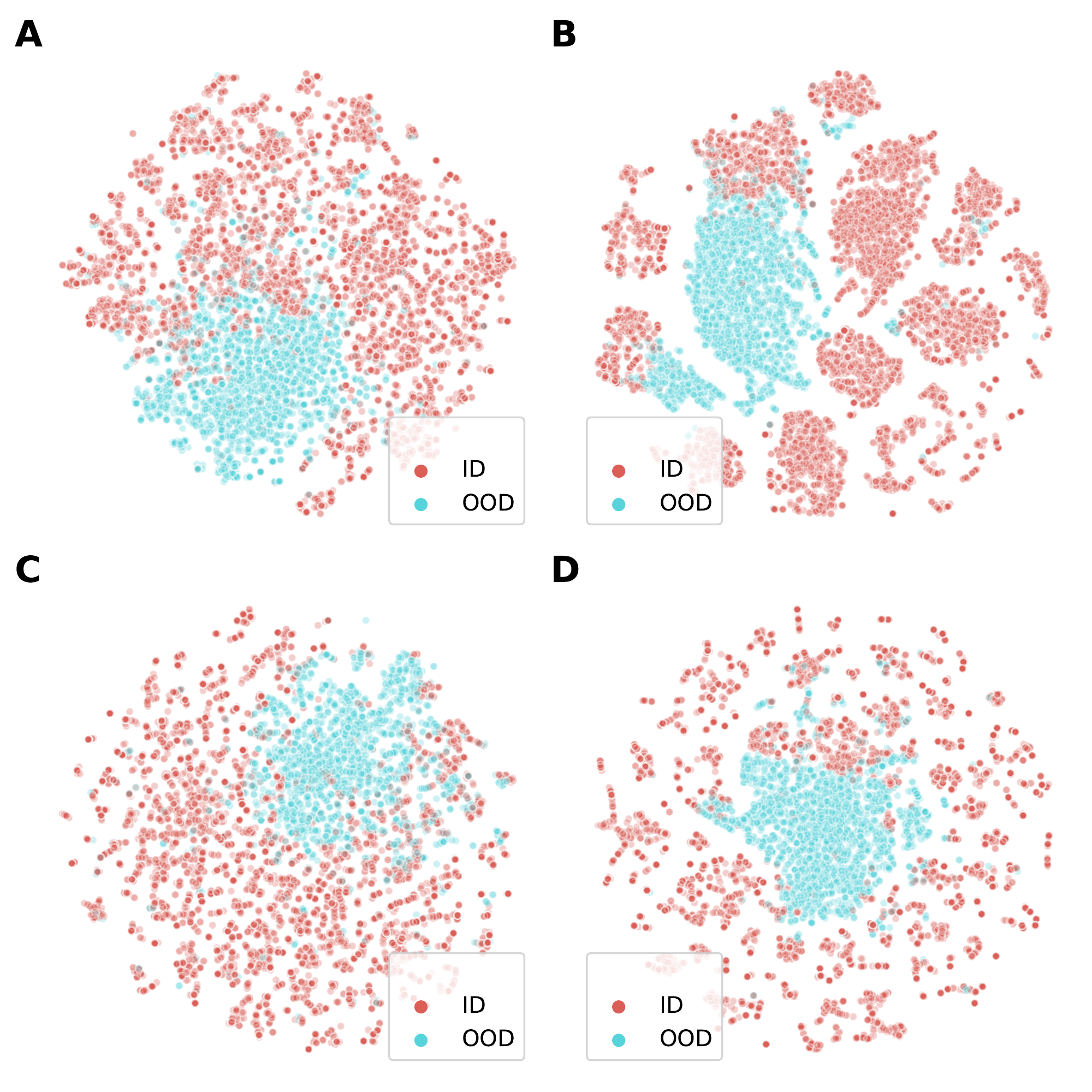}
\caption{T-SNE visualization of learned sentence representations for both ID and OOD samples of ROSTD dataset. A: DistilBERT-Base-NLI; B: DistilBERT-Base-NLI fine-tuned via unsupervised clustering; C: DistilBERT-Base-NLI fine-tuned via contrastive learning; D: DistilBERT-Base-NLI fine-tuned via combining clustering and contrastive learning.}
\label{fig:representation-rostd}
\end{figure}

To obtain a qualitative sense of how our proposed representation learning method works, we provide T-SNE visualization plots of sentence representations for both ID and OOD samples of the Searchsnippets, SNIPS, Stackoverflow, and ROSTD datasets in Figure \ref{fig:representation-searchsnippets}, \ref{fig:representation-snips}, \ref{fig:representation-stackoverflow}, and \ref{fig:representation-rostd}, respectively. As can be seen, the unsupervised clustering module can aggregate data points into many clusters by bringing close sentences of the same classes while pushing away sentences of different classes (e.g., Figure \ref{fig:representation-searchsnippets}B, \ref{fig:representation-snips}B, \ref{fig:representation-stackoverflow}B, and \ref{fig:representation-rostd}B). In contrast, the contrastive learning module can distribute data points uniformly over the whole latent space, in which process ID and OOD samples can be pushed away from each other (e.g., Figure \ref{fig:representation-searchsnippets}C). By combining these two modules, we can realize both goals and make the boundaries between ID and OOD samples much more clear compared with no adaptive fine-tuning (e.g., Figure \ref{fig:representation-searchsnippets}A vs. Figure \ref{fig:representation-searchsnippets}D), and thus achieve the optimal OOD detection performance. 
Note when both modules are combined, the OOD samples tend to be condensed into the center, while ID samples are scattered outside of OOD samples in clusters (e.g., Figure \ref{fig:representation-searchsnippets}D, \ref{fig:representation-snips}D, \ref{fig:representation-stackoverflow}D, and \ref{fig:representation-rostd}D), which facilitates the density estimator to better differentiate ID and OOD samples. 

More specifically, for the Searchsnippets dataset, without the help of contrastive learning, representations obtained via only clustering still cannot well differentiate OOD samples from ID samples because ID and OOD samples still mingle with each other for some clusters (see Figure \ref{fig:representation-searchsnippets}B). In contrast, from Figure \ref{fig:representation-searchsnippets}C, we find that contrastive learning alone can better separate ID and OOD samples than no adaptation by condensing the ID data distribution (red dots) and spreading OOD samples outside (blue dots). In this case, the contrastive learning module contributes more to the full model performance than the clustering module due to its better capability at differentiating ID and OOD samples, which aligns with the conclusion drawn from the ablation study in Table \ref{tab:final-performance-comparison}. However, for the ROSTD and SNIPS datasets, unsupervised clustering itself can already well separate ID and OOD samples by distributing the ID samples into many condensed clusters, where OOD samples are distributed outside of the ID clusters. This explains the finding revealed by the Ablation study in Section \ref{sec:ablation-study} that the clustering objective contributes more to the full model performance. 
% Such representation visualizations for other datasets are summarized in the Appendix \ref{sec:visualization_representations}. 

\section{Conclusion}

In this work, we aim at the OOD detection problem without using any ID labels. 
In addition to evaluating different LM based likelihood methods, we propose a representation learning based method by performing unsupervised clustering and contrastive learning to learn good data representations for OOD detection. We demonstrate that this novel unsupervised method can not only outperform the best likelihood based methods but also be even competitive to the state-of-the-art supervised method that has extensively used labeled data.

\appendices

\setcounter{table}{0}
\setcounter{figure}{0}
\setcounter{footnote}{0}
\renewcommand\thetable{\Alph{section}.\arabic{table}}
\renewcommand\thefigure{\Alph{section}.\arabic{figure}}

% \section{Visualizations of Representations}
% \label{sec:visualization_representations}


% \section{}

% \section*{Acknowledgment}


\ifCLASSOPTIONcaptionsoff
  \newpage
\fi



\bibliographystyle{IEEEtran}
\bibliography{IEEEabrv,references}

% \end{thebibliography}


% \end{IEEEbiography}


%\newpage



%\vfill


\end{document}